# Effective Human Activity Recognition Based on Small Datasets

Bruce X. B. YU
Department of Computing
The Hong Kong Polytechnic University
Hong Kong, China
bruce.xb.yu@connect.polyu.hk

Yan Liu
Department of Computing
The Hong Kong Polytechnic University
Hong Kong, China
csyliu@comp.polyu.edu.hk

Keith C. C. CHAN
Department of Computing
The Hong Kong Polytechnic University
Hong Kong, China
keith.chan@polyu.edu.hk

*Abstract*— Most recent work on vision-based human activity recognition (HAR) focuses on designing complex deep learning models for the task. In so doing, there is a requirement for large datasets to be collected. As acquiring and processing large training datasets are usually very expensive, the problem of how dataset size can be reduced without affecting recognition accuracy has to be tackled. To do so, we propose a HAR method that consists of three steps: (i) data transformation involving the generation of new features based on transforming of raw data, (ii) feature extraction involving the learning of a classifier based on the AdaBoost algorithm and the use of training data consisting of the transformed features, and (iii) parameter determination and pattern recognition involving the determination of parameters based on the features generated in (ii) and the use of the parameters as training data for deep learning algorithms to be used to recognize human activities. Compared to existing approaches, this proposed approach has the advantageous characteristics that it is simple and robust. The proposed approach has been tested with a number of experiments performed on a relatively small real dataset. The experimental results indicate that using the proposed method, human activities can be more accurately recognized even with smaller training data size.

## I. INTRODUCTION

Skeleton based HAR has been an active filed since the release of the motion sensor known as Kinect. With the booming of deep learning (DL) algorithms, researches are collecting larger datasets to feed their DL models. As illustrated in Fig. 1 [1], it has been commonly agreed that the more data will help to deliver the better performance on neural network models. Or alternatively, you need to design a better feature extraction model to improve the performance. Feature extraction involves reducing the required resources for a dataset while still describing the data with sufficient accuracy. When performing analysis of complex data, one major problem stems from the number of variables involved. Analyzing a large number of variables generally requires a large amount of memory and computational resources, also it may render a classification algorithm easy to overfitting and poor in generalization power. Many machine learning practitioners believe that properly optimized feature extraction is the key to effective model construction [2]. For HAR tasks, collecting large volume of activity data is not only intrusive for users but also time consuming for developers. For developers, it will be costly in terms of factors like storage, labeling labor, and computational resource. Due to these feasibility issues, we propose a HAR framework with feature extraction layer that improves the HAR accuracy on a small household HAR dataset.

Recent studies increase the data volume by installing more cameras to different positions like in the data collection setups of [3] and [4], which will need data centralization and calibra-tion during training and evaluation steps. Although multi view solutions with more cameras will be intuitively more capable, but it will encounter issues like hard of installation, poor user acceptance, and low affordability. Furthermore, to the best of our knowledge, existing vision based HAR are all conducted in the laboratory with very fixed setups and there is no solution tackling real world environment. With above concerns, small HAR datasets for household application development has its market and deserves attention as it could be easy for customization. Unlike the complicated arbitrary setups of previous jobs conducted in the laboratory, we proposed a HAR method for in a real-world environment that is agile, acceptable, and affordable. Precisely, we install a Kinect v2 sensor on the ceiling of an elderly person's living room and recognize the daily activities of the elderly subject.

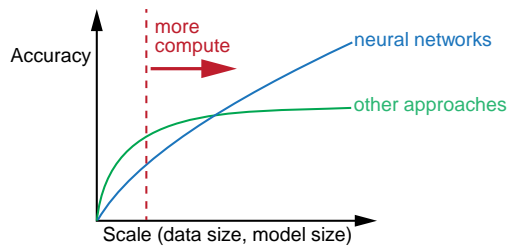

Fig. 1. Relationship of model accuracy and data size

The main contributions of this paper could be fourfold. First, we propose a HAR framework with feature extraction processes. The feature extraction algorithm named ABEF is used to extract temporal features from spatial temporal skeleton activity data. The dimension of the extracted features is much smaller than that of the raw Kinect skeleton data. Meanwhile it maintains the discriminative ability, which makes classifiers easier to separate one activity from the other ones. Second, we collect a real-world scenario dataset in an elderly person's home. One Kinect v2 sensor is used to collect daily activities of an elderly person, which is consistent with the general installation of existing commercial household surveillance cameras. Comparing with existing methods, our data collection method is simple yet capable for HAR tasks. Third, to verify the effectiveness of our HAR framework, we conduct various experiments on the collected dataset by comparing our method with some benchmark DL models. It turns out that our method outperforms the current learning-based algorithms. Lastly, the experimental results highlight the ignored importance of feature extraction for many DL based approaches.

## II. RELATED WORK

In early days, Wang et al. [5] proposed to use the first version of Kinect sensor to recognize human activities like drink, eat, read book, and call cell phone. The Kinect sensor is installed in front of the subjects who perform actions at a fixed location which is a big sofa. Comparing with real world sce-narios, the setup of [5] is quite fixed and the scale of the monitoring area is very limited, and it is easy to be affected by the movement of furniture. Since the angle of view is important for deploying a model to real scenario applications, there are some recent methods using increasing number of cameras to monitor the subjects from various angles. One of recently representative multi-view HAR method that uses three cameras is proposed by [3], which tries to simultaneously classify activities of three categories: daily actions, medical actions and mutual actions. Another similar job is propose by [6] that is even more arbitrary using 8 cameras to surround the monitored subject for designing HAR module of a moving robot. Concerning the feasibility, installing more cameras in application scenarios like household environment might be costly and difficult. The general setup of household surveillance cameras is usually with one camera installed on the corner of the ceiling. For feeding data hungry models with big data, the state-of-the-art DL algorithms has achieved great performance. However, counting the labeling and tuning time, there remains a gap between current HAR solutions and real scenario applications.

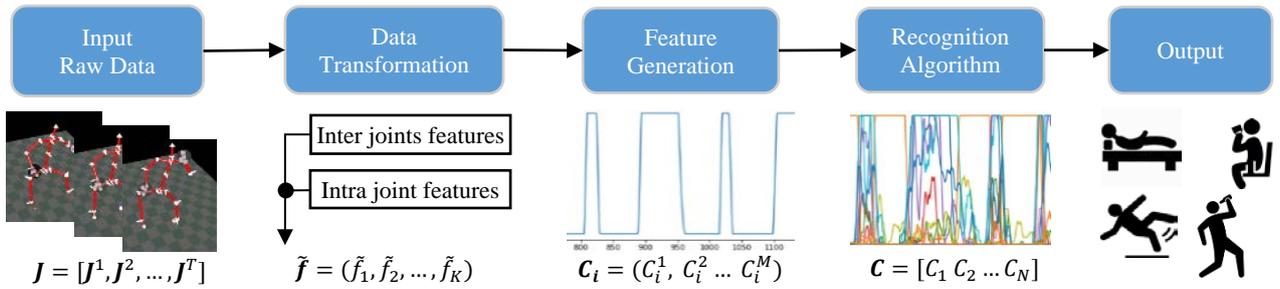

Fig. 2. Our proposed HAR framework with feature extraction

Regarding HAR algorithms, traditionally algorithms like the DTW [7], HMM [8], and SVM [9] have been proposed for developing predictive models for HAR based on skeleton data. More recently, deep learning algorithms [10] have been used for this task. The relative merits of these algorithms depend on such factors as accuracy, processing speed, and ease-of-deployment and there is always a need for us to develop an algorithm that can perform better according to these factors. The advance of deep learning makes it possible to perform automatic high-level feature extraction thus achieves promising performance in many areas. Since then, DL based methods have been widely adopted for skeleton based HAR tasks. Wang et al. [10] reviewed DL models for HAR tasks, which includes Deep neural network, Convolutional Neural Network, Stacked autoencoder etc. Concerning spatial and temporal characteristics of the skeleton data, Shahroudy et al. [11] proposed a Spatial Temporal LSTM (ST-LSTM) algorithm to learning representation from skeleton data. With the similar concern, Yan et al. [12] proposed a Spatial Temporal-Graph Convolutional Network (ST-GCN) that learns both the spatial and temporal patterns from skeleton activity data. Many enhanced versions of ST-GCN models has been proposed by considering other physical prior knowledge. For example, Lei et al. [13] proposed a non-local GCN that leans the graph structure individually for different layers and samples and achieved improved performance than the manually designed graph of Yan. Another GCN method proposed by Li et al. [14] tries to model discriminative features from actional and structural links of the skeleton graph. These DL methods are all rely on big datasets and whether they will work properly when no such big data are available has not been tested. Hence in this job, we claim that feature extraction that is invariant to view changing could be effective for real world HAR scenarios when collecting large training data is not convenient.

III. HAR FRAMEWORK WITH FEATURE EXTRACTION

This section introduces the proposed HAR framework with a feature extraction step that is implemented with an AdaBoost Feature Extraction (ABFE) algorithm. This is unlike most existing jobs that directly feed the whole raw skeleton data to DL models.

*A. HAR Framework*

The raw skeleton data of activities is characterized as multivariate, spatial, and temporal. The spatial relationship among skeleton joints, if properly extracted, will contribute to the separational power of a classifier. Similarly, the sequential patterns of inter joints and intra joint are also the interesting features that need to be mined for improving the classification performance. However, it might be hard to numerating the whole set of features, the job of latent feature learning is to extract the ones that contribute the most to the accuracy. By using DL methods, the raw data could be feed to the model directly and the model will learn neural network weights, which finally delivers features in the final connect layer. However, the features generated by DL models are usually unexplainable and hard to improve and expand to other unnormalized datasets. Hence, feature extraction remains essential for HAR with small training data. In the second step of the proposed framework (see Fig. 2), we adopt an extended version of the AdaBoost algorithm proposed by Rojas [15], which is traditional yet effective for generating discriminative features. In the following subsections, details of each step will be introduced.

*B. Mathematical Formulation of the Skeleton Data*

As discussed above, we used the Kinect v2 sensors to collect our dataset. For any one particular activity being monitored, using the Kinect v2, we record a sequence of skeleton body frames corresponding to the actions performed. Each skeleton body frame consists of 25 joints (see Fig. 3) which can be labelled as HEAD, NECK, ..., FOOTLEFT, etc. For a set of joints in a body frame that is observed at time $t$, let us represent the set as $j^t = (j_1^t, ..., j_i^t, ..., j_{25}^t)$ where $j_i^t$ is the 3-D cartesian coordinates of the position of joint $i$ so that $j_i^t = (j_{ix}^t, j_{iy}^t, j_{iz}^t)$ with $j_{ix}^t$, $j_{iy}^t$, and $j_{iz}^t$ correspond to the values of the x-, y- and z-coordinates, respectively. An activity that begins at time $t = 1$ and ends at time $T$ with body frames collected at regular intervals can, therefore, be represented as a time series of $T$ skeleton frames, $\mathbf{J_i} = [j^1, j^2, ..., j^t, ..., j^T]$.

With $M$ training samples, we will have $\mathbf{J} = \{\mathbf{J}_1, \mathbf{J}_2, \ldots, \mathbf{J}_i, \ldots \mathbf{J}_M\}$ for training.

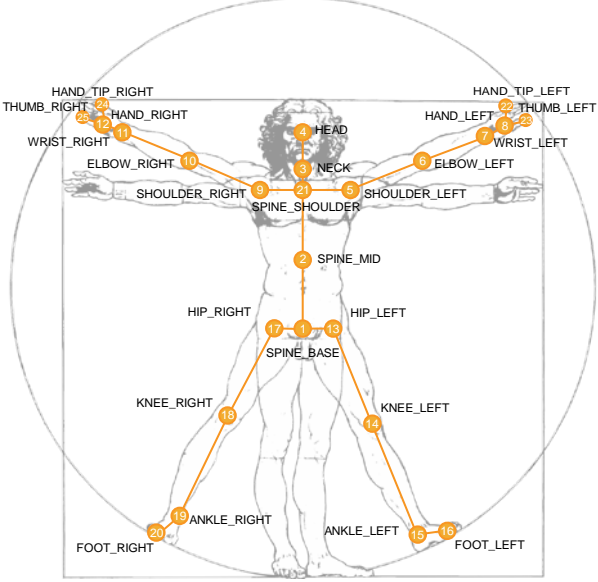

Fig. 3. Skeleton joints of Kinect v2 sensor

*C. Feature Extraction Algorithm*

One HAR algorithm that we can use for the proposed framework is the AdaBoost algorithm [16]. A typical AdaBoost algorithm can be trained to recognize a particular activity by developing a binary classifier. In other words, in the case that there are multiple activities to be recognized, we develop a binary classifier for each of them using the AdaBoost algorithm which we describe as follows.

Given $T$ skeleton joint frames $\mathbf{J}_i = [\mathbf{j}^1, \mathbf{j}^2, \ldots, \mathbf{j}^t, \ldots, \mathbf{j}^T]$, for $\mathbf{j}^t = (\mathbf{j}_1^t, \ldots, \mathbf{j}_i^t, \ldots, \mathbf{j}_{25}^t)$ and $\mathbf{j}_i^t = (j_{ix}^t, j_{iy}^t, j_{iz}^t)$, it will be transformed to both inter and intra joint features with functions $\mathbf{f} = (f_1(\cdot), f_2(\cdot), \ldots, f_i(\cdot), \ldots, f_K(\cdot))$, where $f_i(\cdot)$ is one of $K$ such transformation function for $\mathbf{J}_i$, whose results can then be represented as $\tilde{\mathbf{f}} = (\tilde{f}_1, \tilde{f}_2, \ldots, \tilde{f}_i, \ldots, \tilde{f}_K)$. Some representative functions that are utilized are given in Table I below. For example, the joint position distance is one of the inter joints features that could be calculated as:

$$\tilde{f}_i = f_i(\mathbf{j}_k^t, \mathbf{j}_l^t) \quad (1)$$

$$f_i(\mathbf{j}_k^t, \mathbf{j}_l^t) = \sqrt{(j_{kx}^t - j_{lx}^t)^2 + (j_{ky}^t - j_{ly}^t)^2 + (j_{kz}^t - j_{lz}^t)^2} \quad (2)$$

where $\mathbf{j}_k^t$ and $\mathbf{j}_l^t$ are two joints of the skeleton at time $t$. In the implementation, only the distance of representative joint pairs that considered as effective feature will be used.

TABLE I. EXAMPLES OF FEATURE MODELING FUNCTIONS

| Inter joints features | Intra joint features |
|---|---|
| Joint position distance | Velocity in 3D space |
| Angles between 3 joints | Speed |
| Velocity of angles | Acceleration |
| Acceleration of angles | Muscle force |

In other words, the training data $\mathbf{J} = \{\mathbf{J}_1, \mathbf{J}_2, \ldots, \mathbf{J}_i, \ldots \mathbf{J}_M\}$ can be described in terms of these function as latent features $\tilde{\mathbf{f}} = (\tilde{f}_1, \tilde{f}_2, \ldots, \tilde{f}_i, \ldots, \tilde{f}_K)$. The $\tilde{\mathbf{f}}$ is then used to train a binary classifier for each activity by using the AdaBoost algorithm [15]. Each feature $\tilde{f}_i$ will be used to build a weak classifier $g_i(\tilde{f}_i)$. In the scouting step of AdaBoost, with $K$ weak classifiers, an expert pool which is represented as a matrix will be used to record the misses (with a 1) and hits (with a 0) of each classifier on every sample of the training (see Table II).

TABLE II. WEAK THRESHOLD CLASSIFIERS BASED ON FEATURES

|  | $g_1$ | $g_2$ | $\cdots$ | $g_k$ |
|---|---|---|---|---|
| $\mathbf{J}_1$ | 0 | 1 | $\ldots$ | 1 |
| $\mathbf{J}_2$ | 0 | 0 | $\ldots$ | 1 |
| $\vdots$ | $\vdots$ | $\vdots$ | | $\vdots$ |
| $\mathbf{J}_M$ | 0 | 0 | $\ldots$ | 0 |

Elements in the expert pool matrix will be firstly initialized with weights $w_i = 1/M$. Then all the weights will be optimized by gradient descent. In the m-th iteration of the gradient update loop, the weight $w_i^{m+1}$ will be updated by $\alpha_m = \ln((1 - err_m)/err_m)$ as $w_i^{(m)} e^{\pm \alpha_m}$, where $err_m$ is calculated by equation (3):

$$err_m = \frac{\sum_{i=1}^{M} w_i e^{\alpha_m}}{\sum_{i=1}^{M} w_i} \quad (3)$$

The final strong classifier $G(f)$ in (4) is a sign function of the sum of the top ten features selected from the expert pool.

$$G(f) = sign\left(\sum_{m=1}^{K} \alpha_m g_m(f_m)\right) \quad (4)$$

With $N$ strong classifiers $G = \{G_1, G_2, \ldots, G_N\}$ for $N$ activities, the body frames $\mathbf{J}_i$ of each activity will be fed to the $N$ strong classifiers $G$ to generate a feature matrix $\mathbf{C}_i = [\mathbf{c}_1, \mathbf{c}_2, \ldots, \mathbf{c}_i, \ldots, \mathbf{c}_N]$. Fig. 4 shows visualized views of the feature matrix $\mathbf{C}_i$ and the vector $\mathbf{c}_i$ retrieved from all classifiers $G$ and one specific classifier $G_i$, respectively.

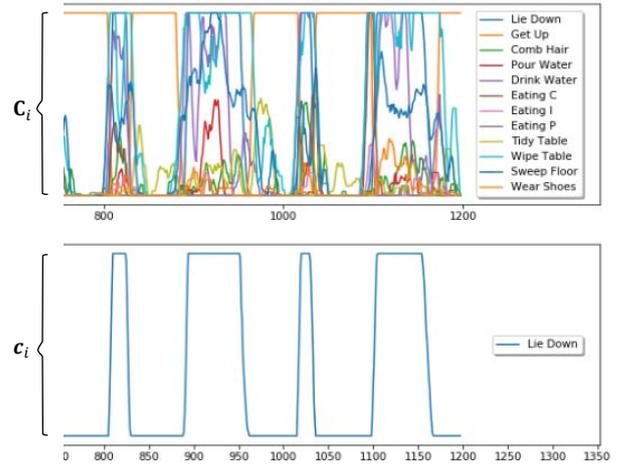

Fig. 4. Visualizd views of the output of ABFE Algorithm for an activity

We name this feature generation method as Adaptive Boosting Feature Extraction (ABFE) and summarized it in Algorithm 1, which is a feature-level method that reduce the dimension of the skeleton frames. The extracted feature matrix $\mathbf{C}_i$, for $\mathbf{c}_i \in \mathbb{R}^T$ could then be fed to different DL models to be trained for inferring its activity.

**Algorithm 1:** ABFE Algorithm

**Data:** $J = \{J_i \mid i = 1, ..., M\}$ dataset for training
**Result:** $C = \{C_i \mid i = 1, ..., M\}$ low dimensional representation of $J$

1. $f = (\tilde{f}_1, \tilde{f}_2, ..., \tilde{f}_i, ..., \tilde{f}_K)$ = features transformed from $J_i$
2. $g_i(\tilde{f}_i)$ = weak classifier in the AdaBoost expert pool
3. N = number of activities
4. **for** $n = 1$ **to** N
5.    $G_n(f)$ = strong binary classifier formed by the top 10 weak classifiers $\{g_i(\tilde{f}_i) \mid i = 1, ..., 10\}$
6. **end**
7. **return** $G = \{G_1, G_2, ..., G_N\}$
8. **for** $i = 1$ **to** $M$ **do**
9.    **for** $j = 1$ **to** $N$ **do**
10.       $c_j$ = feature vector generated by $G_j$
11.    **end**
12.    $C_i = [c_1, c_2, ..., c_N]$ low dimensional representation of $J_i$
13. **end**
14. **return** $C = \{C_i \mid i = 1, ..., M\}$

### D. Recognition Algorithm

Given the extracted low dimensional features $C_i = [c_1, c_2, ..., c_i, ..., c_N]$, $c_i \in \mathbb{R}^T$ as depicted in Fig. 4, we then use it to infer the activity by using an algorithm to represent the multivariate time series features. Specifically, we adopt the Multivariate Long Short-Term Memory Fully Convolutional Network (MLSTM-FCN) algorithm proposed in [17]. The algorithm is built upon the long short-term memory (LSTM) RNNs that is capable to learn temporal dependencies and the Convolutional Network. The LSTM modules is depicted by Graves [18] as:

$$g^c = \sigma(W^c h_{t-1} + I^c c_t) \quad (5)$$

$$g^o = \sigma(W^o h_{t-1} + I^o c_t) \quad (6)$$

$$g^f = \sigma(W^f h_{t-1} + I^f c_t) \quad (7)$$

$$g^u = \sigma(W^u h_{t-1} + I^u c_t) \quad (8)$$

$$m_t = g^f \odot m_{t-1} + g^u \odot g^c \quad (9)$$

$$h_t = tanh(g^o \odot m_t) \quad (10)$$

where $g^c, g^o, g^f, g^u$ are the activation vectors of cell state, output, forget and input gates, respectively. The recurrent weight matrices are denoted by $W^c, W^o, W^f$ and $W^u$. The projection matrices are represented as $I^c, I^o, I^f, I^u$. While $h_t$ is the hidden state vector of the LSTM unit, σ is the logistic sigmoid function, and $\odot$ is the elementwise multiplication. On top of the LSTM unites, an attention mechanism that is a context vector depending on a sequence of annotations $(b_1, ..., b_{T_c})$, where $T_c$ is the maximum length of the sequence $c$. While the FCN module has a squeeze-and-excitation block that performs as will lead to the output for a single dimension as:

$$\tilde{c}_d = F_{scale}(u_d, s_d) \quad (11)$$

where $\tilde{C} = [\tilde{c}_1, ..., \tilde{c}_N]$, $u_d$ is the squeezed feature map generated by a channel-wise global average pooling, $s_d$ is the excitation feature calculated from $u_d$ by a sigmoid function followed by a ReLU function, and $F_{scale}(u_d, s_d)$ denotes the channel wise multiplication of $u_d$ and $s_d$.

## IV. EXPERIMENTS

We now fully describe the data collection method, which is followed by experimental results and discussion.

### A. Data Collection

Thanks to the capability of Kinect V2 sensor, collecting a human activity is quite convenient once the sensor in installed on the ceiling of an elderly person's bedroom. The RGB view of our sensor setup is as depicted in Fig. 5, which cover the whole area of the bedroom from three dimensions namely, height, width and depth. The goal of our setup is to make the HAR solution simple yet effective for real world HAR applications scenarios. This installation method is the same as commercial cameras, which is ideal for household application scenarios like abnormal activity detection, fall detection, daily routine collection, etc. The average frame of the segmented dataset on the RGB stream is 35.6 frames. While the shortest and longest ones last for 6 and 73 frames, respectively.

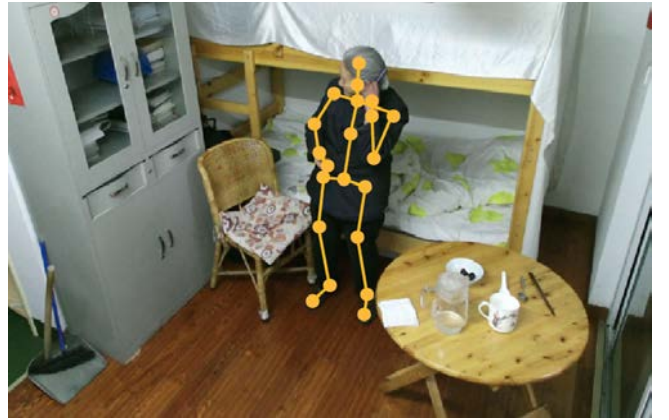

Fig. 5. Real world HAR environment in an edlerly's bedroom

After the raw data are collected, the segmentation part is the main time-consuming work. We developed a program to retrieve the skeleton channel and the RGB channel from the raw data and save them as sequential data in MongoDB and hard disk, respectively. The numbering of different channels in the Kinect v2 raw data is not consistent, which means for one activity there might have 100 RGB frames, 90 skeleton frames and 70 IR frames. Hence, we follow the Kinect v2 raw data frames where all channels are available, otherwise we just drop the Kinect v2 raw frames. As we keep the frames from RGB channel and skeleton channel consistent, we label the segmentation boundaries according to the visual RGB channel and use the segmentation boundaries to segment the skeleton frames. Table III lists the activity classes and their corresponding number of performing times (denoted as #).

TABLE III. DAILY ACTIVITY CLASSES IN OUR DATASET

| Activity Class | Activity Name | # |
|---|---|---|
| 01 | Lie down | 9 |
| 02 | Get up | 9 |
| 03 | Comb hair | 11 |
| 04 | Pour water | 9 |
| 05 | Drink water | 9 |
| 06 | Eat with chopsticks | 10 |
| 07 | Eat with iron spoon | 10 |
| 08 | Eat with pottery spoon | 12 |
| 09 | Tidy table | 16 |
| 10 | Wipe table | 9 |
| 11 | Sweep the floor | 19 |
| 12 | Wear shoes | 17 |

## B. Evaluation Metric

To make the validation less biased than simply splitting the data to training set and test set, we adopt k-fold cross-validation evaluation method with k set to 5 that follows the tradition to make the division of training and test sets representative for measuring the fit of our model based on the volume of the collected dataset. Cross-validation is popularly used to estimate the skill of a machine learning model when the training data is not large [19]. We use top-1 accuracy as in equation (12) which means the prediction must be the same as the ground truth label as the evaluation metric for classification tasks.

$$P = \frac{1}{N}\sum_{k=1}^{N} result_k = \begin{cases} 1 & if\ output_k^{top-1}=label_k \\ 0 & otherwise \end{cases} \quad (12)$$

With the accuracy measure set as top-1, a confusion matrix, also known as error matrix, can be constructed as shown in Table IV. This matrix could be used to visualize the performance of a supervised classification algorithm with two or more classes [20]. In other words, the accuracy of an HAR prediction model can be gauged from the matrix. From the confusion matrix, we can derive the precision and recall measure of each class of activity. For better comparison of classification accuracy, a normalized confusion matrix could also be used. In our experiments, for each confusion matrix corresponding to an activity, we accumulate five folds and normalize the entries in the confusion matrix for further comparison of the overall performances of different HAR algorithms.

TABLE IV. CONFUSION MATRIX WITH N CLASSES

| Actual Class | 1 | $n_{11}$ | $n_{12}$ | … | $n_{1N}$ |
|---|---|---|---|---|---|
| | 2 | $n_{21}$ | $n_{22}$ | … | $n_{2N}$ |
| | ⋮ | ⋮ | ⋮ | … | ⋮ |
| | N | $n_{N1}$ | $n_{N2}$ | … | $n_{NN}$ |
| | | 1 | 2 | … | N |
| | | **Predicted Class** | | | |

## C. Experimental Settings

We compare other two representative DL models mentioned before to justify the effectiveness of our proposed feature extraction step. The first model is ST-GCN [12] and the second model is ST-LSTM [11]. For ST-GCN, we keep the original implementation of the author and change the number of the output to 12. Since we use the Kinect v2, the joints number is set as 25. The number of people is 1 as our dataset only involves one subject. The maximum sequential frame length is set to 75 based on the statistics of the collected dataset. Hence, for one activity, the input to the ST-GCN in our experiment is a tensor with shape (3,25,75,1). Empirically, we set the initial learning rate is 0.1 and will decay to 1/10 of the precious learning rate at epochs of 40, 100, and 150. The predication result will be evaluated with an interval of 10 epochs. We train the model with a batch size of 64 and terminate the training at epoch 200 and show the best result throughout the training process. For ST-LSTM, we empirically follow the original hyper parameter setting except change the sequence length from 6 to 10, set the evaluation interval to 10, and terminate the training at epoch 200. The best result from all evaluations is selected to show in the experimental results.

In our method, the MLSTM-FCN model [17] that comprises of a fully convolutional block and a LSTM block that perform as feature extractors and finally concatenated together to a SoftMax layer is implemented by following the setting on the Arabic Voice dataset which is similar with the characteristic of our extracted features from the ABFE. We set the batch size and total epoch the same as the compared two DL models and show the best results in the next section.

By using the cross-validation method, both the skeleton dataset *J* and its transformed form as *C* are divided into five folds. Once one of the cross-validation folds is selected as testing set, the other three folds are used to train the model. There is no sample duplicated among the cross-validation folds. All the experiments are implemented on a Supermicro GPU Server (model SYS-7048GR-TR) with 4 GTX 1080 Ti GPUs. All the GPUs are used in each experiment.

## D. Experimental Results

The experimental results for ST-GCN, LSTM and our method are listed in Table V, which indicates the top-1 accuracy of each cross-validation fold and their average accuracy. To investigate the training speed, we recorded the starting time and ending time, then calculated the differences of them. Table VI gives the training time of all cross-validation folds. Specifically, for ST-GCN and ST-LSTM, the starting time and ending time are based on the generation time of the first checkpoint (at the epoch 10) and the last checkpoint (both at the epoch 200), respectively. For our method, we recorded the starting time and ending time of each training process.

TABLE V. TOP-1 ACCURACY ON VARIOUS MODELS

| Algorithm | Fold 1 | Fold 2 | Fold 3 | Fold 4 | Fold 5 | Average |
|---|---|---|---|---|---|---|
| *ST-LSTM* | 76.67 | 79.31 | 75.00 | 75.86 | 87.50 | 78.87 |
| *ST-GCN* | 66.67 | 77.41 | 67.86 | 55.17 | 75.00 | 68.42 |
| *Our Method* | **92.86** | **92.59** | **96.15** | **96.15** | **96.43** | **94.84** |

TABLE VI. TRAINING TIME OF ALL EXPERIMENTAL SETS

| Algorithm | Fold 1 | Fold 2 | Fold 3 | Fold 4 | Fold 5 | Average |
|---|---|---|---|---|---|---|
| *ST-LSTM* | 0:24:12 | 0:25:26 | 0:28:24 | 0:24:29 | 0:30:35 | 0:26:37 |
| *ST-GCN* | 0:09:38 | 0:09:49 | 0:09:43 | 0:09:48 | 0:09:11 | 0:09:38 |
| *Our Method* | **0:00:55** | **0:00:53** | **0:00:51** | **0:00:53** | **0:00:51** | **0:00:53** |

Except showing the top-1 accuracy and the training time, to further ease the comparison of the implemented methods, we show the accumulated and normalized confusion matrices of all three algorithms in Table VII, VIII, and IX.

## E. Effectiveness Evaluation

According to the experimental results of Table V, the extracted feature matrix from our proposed ABFE algorithm achieved the highest top-1 accuracy in all five cross-validation folds. The average accuracy with the value of 94.84% is significantly better than that of data driven methods. From the results of ST-GCN and ST-LSTM in their cross-validation folds, we could observe that data driven methods requires the samples in the test set to have at least similar samples in the training sample to achieve the recognition or overfitting. It indicates that our feature extraction method successfully reduces the data size and maintains and even surpass the discriminative power of the skeleton data.

TABLE VII. ACCUMULATED AND NORMALIZED CONFUSION MATRIX OF ST-LSTM

| True Label | 1 | 2 | 3 | 4 | 5 | 6 | 7 | 8 | 9 | 10 | 11 | 12 |
|---|---|---|---|---|---|---|---|---|---|---|---|---|
| 1 | 33.3% | 33.3% | 0.0% | 0.0% | 22.2% | 0.0% | 0.0% | 0.0% | 0.0% | 0.0% | 0.0% | 11.1% |
| 2 | 0.0% | 100.0% | 0.0% | 0.0% | 0.0% | 0.0% | 0.0% | 0.0% | 0.0% | 0.0% | 0.0% | 0.0% |
| 3 | 0.0% | 0.0% | 100.0% | 0.0% | 0.0% | 0.0% | 0.0% | 0.0% | 0.0% | 0.0% | 0.0% | 0.0% |
| 4 | 0.0% | 0.0% | 0.0% | 77.8% | 22.2% | 0.0% | 0.0% | 0.0% | 0.0% | 0.0% | 0.0% | 0.0% |
| 5 | 0.0% | 0.0% | 0.0% | 11.1% | 77.8% | 0.0% | 11.1% | 0.0% | 0.0% | 0.0% | 0.0% | 0.0% |
| 6 | 0.0% | 0.0% | 0.0% | 0.0% | 0.0% | 20.0% | 80.0% | 0.0% | 0.0% | 0.0% | 0.0% | 0.0% |
| 7 | 0.0% | 0.0% | 0.0% | 0.0% | 0.0% | 0.0% | 50.0% | 50.0% | 0.0% | 0.0% | 0.0% | 0.0% |
| 8 | 0.0% | 0.0% | 0.0% | 0.0% | 0.0% | 0.0% | 41.7% | 58.3% | 0.0% | 0.0% | 0.0% | 0.0% |
| 9 | 0.0% | 0.0% | 0.0% | 0.0% | 0.0% | 0.0% | 0.0% | 0.0% | 100.0% | 0.0% | 0.0% | 0.0% |
| 10 | 0.0% | 0.0% | 0.0% | 0.0% | 0.0% | 0.0% | 0.0% | 0.0% | 0.0% | 100.0% | 0.0% | 0.0% |
| 11 | 0.0% | 0.0% | 0.0% | 0.0% | 0.0% | 0.0% | 0.0% | 0.0% | 0.0% | 0.0% | 100.0% | 0.0% |
| 12 | 0.0% | 0.0% | 0.0% | 0.0% | 0.0% | 0.0% | 0.0% | 0.0% | 0.0% | 0.0% | 0.0% | 100.0% |

Predicted Label

TABLE VIII. ACCUMULATED AND NORMALIZED CONFUSION MATRIX OF ST-GCN

| True Label | 1 | 2 | 3 | 4 | 5 | 6 | 7 | 8 | 9 | 10 | 11 | 12 |
|---|---|---|---|---|---|---|---|---|---|---|---|---|
| 1 | 77.8% | 22.2% | 0.0% | 0.0% | 0.0% | 0.0% | 0.0% | 0.0% | 0.0% | 0.0% | 0.0% | 0.0% |
| 2 | 66.7% | 33.3% | 0.0% | 0.0% | 0.0% | 0.0% | 0.0% | 0.0% | 0.0% | 0.0% | 0.0% | 0.0% |
| 3 | 0.0% | 0.0% | 100.0% | 0.0% | 0.0% | 0.0% | 0.0% | 0.0% | 0.0% | 0.0% | 0.0% | 0.0% |
| 4 | 0.0% | 0.0% | 0.0% | 55.6% | 0.0% | 0.0% | 0.0% | 0.0% | 0.0% | 0.0% | 0.0% | 44.4% |
| 5 | 0.0% | 0.0% | 0.0% | 22.2% | 66.7% | 0.0% | 0.0% | 0.0% | 0.0% | 0.0% | 0.0% | 11.1% |
| 6 | 0.0% | 0.0% | 0.0% | 40.0% | 0.0% | 10.0% | 10.0% | 40.0% | 0.0% | 0.0% | 0.0% | 0.0% |
| 7 | 0.0% | 0.0% | 0.0% | 50.0% | 0.0% | 0.0% | 20.0% | 30.0% | 0.0% | 0.0% | 0.0% | 0.0% |
| 8 | 0.0% | 0.0% | 0.0% | 16.7% | 0.0% | 8.3% | 25.0% | 50.0% | 0.0% | 0.0% | 0.0% | 0.0% |
| 9 | 0.0% | 0.0% | 0.0% | 0.0% | 0.0% | 0.0% | 0.0% | 0.0% | 100.0% | 0.0% | 0.0% | 0.0% |
| 10 | 0.0% | 0.0% | 0.0% | 0.0% | 0.0% | 0.0% | 0.0% | 0.0% | 11.1% | 88.9% | 0.0% | 0.0% |
| 11 | 0.0% | 0.0% | 0.0% | 0.0% | 0.0% | 0.0% | 0.0% | 0.0% | 36.8% | 0.0% | 63.2% | 0.0% |
| 12 | 0.0% | 0.0% | 0.0% | 0.0% | 0.0% | 0.0% | 0.0% | 0.0% | 0.0% | 0.0% | 0.0% | 100.0% |

Predicted Label

TABLE IX. ACCUMULATED AND NORMALIZED CONFUSION MATRIX OF OUR METHOD

| True Label | 1 | 2 | 3 | 4 | 5 | 6 | 7 | 8 | 9 | 10 | 11 | 12 |
|---|---|---|---|---|---|---|---|---|---|---|---|---|
| 1 | 100.0% | 0.0% | 0.0% | 0.0% | 0.0% | 0.0% | 0.0% | 0.0% | 0.0% | 0.0% | 0.0% | 0.0% |
| 2 | 0.0% | 88.9% | 0.0% | 0.0% | 0.0% | 0.0% | 0.0% | 0.0% | 0.0% | 0.0% | 11.1% | 0.0% |
| 3 | 0.0% | 0.0% | 100.0% | 0.0% | 0.0% | 0.0% | 0.0% | 0.0% | 0.0% | 0.0% | 0.0% | 0.0% |
| 4 | 0.0% | 0.0% | 0.0% | 100.0% | 0.0% | 0.0% | 0.0% | 0.0% | 0.0% | 0.0% | 0.0% | 0.0% |
| 5 | 0.0% | 0.0% | 0.0% | 0.0% | 88.9% | 0.0% | 11.1% | 0.0% | 0.0% | 0.0% | 0.0% | 0.0% |
| 6 | 0.0% | 0.0% | 0.0% | 0.0% | 0.0% | 100.0% | 0.0% | 0.0% | 0.0% | 0.0% | 0.0% | 0.0% |
| 7 | 0.0% | 0.0% | 0.0% | 0.0% | 0.0% | 0.0% | 100.0% | 0.0% | 0.0% | 0.0% | 0.0% | 0.0% |
| 8 | 0.0% | 0.0% | 0.0% | 0.0% | 0.0% | 10.0% | 0.0% | 90.0% | 0.0% | 0.0% | 0.0% | 0.0% |
| 9 | 0.0% | 0.0% | 0.0% | 0.0% | 0.0% | 0.0% | 0.0% | 0.0% | 100.0% | 0.0% | 0.0% | 0.0% |
| 10 | 0.0% | 0.0% | 0.0% | 0.0% | 0.0% | 0.0% | 0.0% | 0.0% | 0.0% | 100.0% | 0.0% | 0.0% |
| 11 | 5.9% | 0.0% | 0.0% | 0.0% | 0.0% | 0.0% | 0.0% | 0.0% | 0.0% | 11.8% | 82.4% | 0.0% |
| 12 | 0.0% | 0.0% | 0.0% | 0.0% | 0.0% | 0.0% | 0.0% | 0.0% | 0.0% | 0.0% | 10.0% | 90.0% |

Predicted Label

From the confusion matrix, we could have a closer look at which activities are failed and which activities are easy to be recognized by different algorithms. The figures in Table VII and VIII show that activity 6 (eat with chopsticks) and activity 7 (eat with iron spoon) are the most challenging ones for the tested data driven models. While the feature of these two relatively fine-grained activities is successfully captured by our method as shown in Table IX that indicate zero failure throughout the 5 cross-validation folds. The ST-GCN has the ability to automatically learning both the spatial and temporal patterns from big dataset as in [3] and [4]. However, when encountered with some fine-grained challenging activities like pour water, drink water, and eat with pottery spoon, both spatial temporal DL models failed to learn effective feature throughout all the cross-validation folds.

We also observed from the confusion matrices that different models have their advantage to recognize specific activities. For example, ST-LSTM is the best in recognize activity 2 (get up), ST-GCN and ST-LSTM are both good at recognizing activity 12 (wear shoes), while our method is good at more activities like activity 1 (lie down), activity 4 (pour water), and activity 6 (eat with chopsticks). When the number of activities increases in this job, we could solve this issue by grouping activities according to different locations. In such a way, we could decrease the dimension of feature matrices from the feature generation step.

### F. Efficiency Evaluation

Our proposed method also achieved the shortest average training time (53 seconds for 200 epochs) as shown in Table VI. The training time is significantly less than that of ST-GCN and ST-LSTM that spend around 10 and 25 times more training time than our method, respectively. It indicates that the proposed feature extraction algorithm successfully reduces the data size to a feature matrix. This could be the reason why it has shorter training time than that of the raw skeleton based methods, which makes the training process more efficient than the others and could be an advantage to speed up deployment of real-world solutions.

## V. CONCLUSION AND FUTURE WORK

This paper proposed a HAR framework with a feature transformation method and a feature extraction algorithm ABFE. We model the intra joint and inter joints features and train the transformed feature with AdaBoost to generate feature matrices in the feature extraction step. We then conducted three sets of experiment and found that our proposed method achieved the best performance in terms of effectiveness and efficiency on the real-world dataset that we collected. The experimental results validated the contribution of the proposed HAR framework for developing household HAR applications.

In the future, we will improve the proposed framework by designing a location-based model that encodes other contextual information from other data modalities to narrow the searching space of the model. Instead of arbitrarily increasing the data volume and feeding them to DL models, with this HAR framework, our future jobs will make HAR solutions more explainable and improvable. Although the classification step in our framework is not in real-time, the other steps of the proposed framework are ready for online processing, hence we will also design online HAR solutions in the future.